\documentclass{article}
\usepackage{ijcai19}

\usepackage{times}
\usepackage{soul}
\usepackage{url}
\usepackage[hidelinks]{hyperref}
\usepackage[utf8]{inputenc}
\usepackage[small]{caption}
\usepackage{graphicx}
\usepackage{amsmath}
\usepackage{booktabs}
\usepackage{algorithm}
\usepackage{algorithmic}
\usepackage{multirow}
\usepackage{xcolor,graphicx,float}
\usepackage{subfigure}
\usepackage{multirow}

\title{A Learning based Branch and Bound \\for Maximum Common Subgraph Problems}

\author{
Yan-li Liu$^{1,3}$\and
Chu-min Li$^2$\and
Hua Jiang$^{2}$\footnote{Contact Author}\and
Kun He$^1$ \footnote{Contact Author}\\
\affiliations
$^1$Huazhong university of science and technology, China\\
$^2$University of Picardie Jules Verne, France\\
$^3$Wuhan university of science and technology, China
\emails
\{yanlil2008,jh\_hgt\}@163.com,
chu-min.li@u-picardie.fr,
Brooklet60@hust.edu.cn
}

\begin{document}

\maketitle

\begin{abstract}
Branch-and-bound (BnB) algorithms are widely used to solve combinatorial problems, and the performance crucially depends on its branching heuristic.
In this work, we consider a typical problem of maximum common subgraph (MCS), and propose a branching heuristic inspired from reinforcement learning with a goal of reaching a tree leaf as early as possible to greatly reduce the search tree size.
Extensive experiments show that our method is beneficial and outperforms current best BnB algorithm for the MCS.
\end{abstract}

\section{Introduction}
A graph is a logic model to describe a set of objects and the relationship of the objects abstracted from real-world applications. Given two graphs $G$ and $H$, it is often crucial to determine the similarities or differences of $G$ and $H$. For this purpose, one should find a graph with as many vertices as possible that is isomorphic to an induced subgraph in both $G$ and $H$. This problem is called {\em Maximum Common induced Subgraph} (MCS). It is NP-hard and widely occurs in applications such as image or video analysis~\cite{Bunke1995Efficient,Liu2001Graph}, information retrieval~\cite{Cao2011Privacy}, biochemistry~\cite{Rosalba2013GRAPES,Adrian2007The,Faccioli2005From}, and pattern recognition~ \cite{Solnon2015On,D2008THIRTY,Cordella2004A}.

Due to the importance of the MCS, many approaches have been designed to solve it, see, e.g, ~\cite{Solnon2010AllDifferent,Mccreesh2016Clique,Vismara2008Finding,Mcgregor2010Backtrack,BahienseEtAl2012,Levi1973,HoffmannEtAl2017}. In this paper, we focus on the branch-and-bound (BnB) scheme for the MCS.
Examples of existing BnB algorithms for the MCS can be found in~\cite{Mcgregor2010Backtrack,NdiayeSolnon2011,Mccreesh2017partition}.

It is well-known that the performance of a BnB algorithm crucially depends on its branching heuristic. The branching heuristic in McSplit~\cite{Mccreesh2017partition}, the best BnB algorithm for the MCS to our knowledge, aims at minimizing the number of branches for the current branching point and uses vertex degrees to rank vertices. Unfortunately, due to the NP-hardness of the MCS, the BnB search tree size may not be predictable in general using vertex degrees or any other superficial feature of a graph.

In this paper, we propose to use reinforcement learning to discover which branching choice yields the greatest reduction of the search tree by trying them out during the search. Concretely, we consider the BnB algorithm as an agent and each branching choice as an action. When the agent takes an action, it receives a reward determined by the consequence of this action, which in our context is the reduction of the search space. The score of the action depends on the accumulated rewards it received in the past. Then, at every branching point, the agent selects, among the actions resulting in the minimum number of branches, the action with the greatest score to branch on, ties being broken in favor of the vertex with the maximum degree.

We implemented our branching heuristic on top of McSplit. The new algorithm is called McSplit+RL and is extensively evaluated on more than 21,743 MCS instances from diverse applications (biochemical reaction, images analysis, 2D, 3D, 4D objects, complex networks), including the large instances used in~\cite{Mccreesh2017partition} for evaluating McSplit. Empirical results show that McSplit+RL solves 130 instances more than McSplit on large graphs, illustrating the effectiveness of combining reinforcement learning in designing branching heuristic for the BnB search.

This paper is organized as follows. Section 2 defines some notations and the MCS problem. Section 3 presents the general BnB scheme for the MCS. Section 4 presents our branching heuristic inspired by reinforcement learning. Section 5 empirically compares McSplit and McSplit+RL and gives insight into the effectiveness of the new branching heuristic. Section 6 concludes.

\section{Problem Definition}

For a simple (unweighted, undirected), labelled graph $G = (V,E,\lambda)$, $V$ is a finite set of vertices, $E\subseteq V\times V$ is a set of edges, and $\lambda$ is a label function that assigns, to each vertex $v \in V$, a label value $\lambda(v)$. If the labels are the same for all vertices, then the labelled graph is reduced to an unlabelled graph. Two vertices $u$ and $v$ are adjacent iff $(u, v) \in E$. The degree of a vertex $v$ is the number of  its adjacent vertices.A subset $V' \subseteq V$ induces a subgraph $G[V'] = (V',E',\lambda')$ of $G$,  where $E'=\{(u, v) \in E | u, v \in V'\}$, and $\forall v \in V'$, $\lambda'(v) = \lambda(v)$.

Given a pattern graph $G_p = (V_p, E_p, \lambda_p)$ and a target graph $G_t = (V_t, E_t, \lambda_t)$, the Maximum Common induced Subgraph (MCS) problem is to find a subset $V'_p \in V_p$ and a subset $V'_t \in V_t$ of the greatest cardinality and a bijection $\phi: V'_p \rightarrow V'_t$ such that: (1) $|V'_p| = |V'_t|$, (2) $\forall v \in V'_p$, $\lambda_p(v) = \lambda_t(\phi(v))$, and (3) given any $v, v' \in V'_p$,  $v$ and $v'$ are adjacent in $G_p$ if and only if $\phi(v)$ and $\phi(v')$ are adjacent in $G_t$. In other words, the MCS is to find a maximum subgraph $G_p[V'_p]$ and a maximum subgraph $G_t[V'_t]$ such that $G[V'_p]$ and $G_t[V'_t]$ are isomorphic. $G_p[V'_p]$ or $G_t[V'_t]$ is called a {\em maximum common induced subgraph} of $G_p$ and $G_t$. The vertex pair $(v, \phi(v))$ is called a {\em match}.

Let $V_p'=\{v_1, v_2, \ldots, v_{|V'_p|}\}$, an optimal solution of the MCS is denoted as a set of matches $\{(v_1, \phi(v_1)), (v_2, \phi(v_2)), \ldots, (v_{|V'_p|}, \phi(v_{|V'_p|})\}$.

\section{Branch and Bound for the MCS}

Given two graphs $G_p$ and $G_t$, the BnB algorithm depicted in Algorithm \ref{alg:Search} gradually constructs and proves an optimal solution using a depth-first search. During the search, the algorithm maintains two variables: $curSol$, the solution under construction; and $maxSol$, the best solution found so far. In addition, every vertex $v$ of $G_p$ is associated with a subset $\alpha(v)$ of vertices of $G_t$ that can be matched with $v$.
At the beginning, $curSol$ and $maxSol$ are initialized to $\emptyset$, and $\alpha(v)$ is initialized to be the set of all vertices of $G_t$ having same label of $v$. The first call MCS$(G_p, G_t, \alpha, \emptyset, \emptyset)$ returns a maximum common subgraph of $G_p$ and $G_t$.

At each branching point, the algorithm first computes an upper bound UB of the cardinality of the best solution that can be found from this branching point, by calling the overestimate$(G_p, G_t, \alpha)$ function. It then compares UB with $maxSol$. If UB $\leq |maxSol|$, a solution better than $maxSol$ cannot be found from this branching point, and the algorithm prunes the current branch and backtracks. Otherwise, it selects a not-yet matched vertex $v$ from $G_p$ and tries to match it with every vertex $w$ in $\alpha(v)$ in turn. As a consequence of matching $v$ with $w$, $(v, w)$ is added into $curSol$, and for each not-yet matched $v'$ of $G_p$, $\alpha(v')$ is updated as follows: If $v'$ is adjacent to $v$, remove all vertices non-adjacent to $w$ from $\alpha(v')$; otherwise, remove all vertices adjacent to $w$ from $\alpha(v')$.

Note that after updating $\alpha(v')$, $\forall w'\in \alpha(v')$, $v$ is adjacent to $v'$ in $G_p$ iff $w$ is adjacent to $w'$ in $G_t$,  so that the match $(v', w')$ can be further added into $curSol$ to yield a feasible solution. If a solution better than $maxSol$ is found in a leaf of the search tree where further match is impossible, the algorithm updates $maxSol$ to be $curSol$ before backtracking to find an even better solution.

An important issue for implementing Algorithm \ref{alg:Search} is how to implement $\alpha$, which determines how to select the vertex $v$ in line \ref{branchingH} and how to design the  overestimate$(G_p, G_t, \alpha)$ function. A natural way is to explicitly create a list of vertices of $G_t$ for each vertex $v$ of $G_p$. With this implementation, Ndiaye and Solnon \citeyear{NdiayeSolnon2011} represent each vertex of $G_p$ as a variable whose domain is a set of vertices of $G_t$. Then, they select the vertex with the smallest domain in line \ref{branchingH}; and use a soft all-different constraint in the overestimate$(G_p, G_t, \alpha)$ function to compute a bound. The difficulty in this implementation of $\alpha$ is that given a vertex $w$ of $G_t$, it is not straightforward to know the number of variables whose domain contains $w$. If the domain of a vertex $v$ is the smallest and contains a vertex $w$ of $G_t$, but $w$ also occurs in the domain of many other vertices of $G_p$, branching on $v$ may not be the best choice to minimize the search tree size.

\begin{algorithm}[H]
\caption{MCS$(G_p, G_t, \alpha, curSol, maxSol)$}
\label{alg:Search}
\textbf{Input}: $G_p=(V_p, E_p, \lambda_p)$, the pattern graph; $G_t=(V_t, E_t, \lambda_t)$, the target graph; $\alpha$, a mapping $V_p \mapsto 2^{V_t}$; $curSol$, the solution under construction; $maxSol$, the best solution found so far\\
\textbf{Output}:  $maxSol$
\begin{algorithmic}[1] 
   \IF { $\forall v\in V_p$, $\alpha(v)$ is empty}
      \IF {$|curSol| > |maxSol|$}
          \STATE $maxSol \leftarrow curSol$;
       \ENDIF
      \STATE \textbf{return}  $maxSol$;
   \ENDIF
   \STATE $\mathit{UB}$ $\leftarrow$ $|curSol|$ + overestimate$(G_p, G_t, \alpha)$; \\
   \IF {$UB$ $\leq |maxSol|$} \label{prune}
      \STATE \textbf{return} $maxSol$;
   \ENDIF
   \STATE $v \leftarrow$ a vertex from $G_p$ such that $\alpha(v) \neq \emptyset$; \label{branchingH}\\
   \FOR {each vertex $w$ in $\alpha(v)$} \label{branchingH1}
       \STATE $\alpha' \leftarrow \alpha $; $\alpha'(v) \leftarrow \emptyset$; \\
       \FOR {each vertex $v'$ in $G_p$} \label{updateD1}
       	    \STATE remove $w$ from $\alpha'(v')$;\\
            \IF {$v'$ is adjacent to $v$ in $G_p$}
                \STATE remove the vertices non-adjacent to $w$ in $G_t$ from $\alpha'(v')$; \\
            \ELSE
               \STATE remove the vertices adjacent to $w$ in $G_t$ from $\alpha'(v')$; \\
            \ENDIF \label{endUpdateVarphi}
       \ENDFOR \label{updateD2}
       \STATE $maxSol \leftarrow$ MCS$(G_p, G_t, \alpha',$\\$~~~~~~~~~~~~~~~~~~~~~ curSol \cup \{(v, w)\}, maxSol)$; \\
   \ENDFOR
   \STATE $\alpha(v) \leftarrow \emptyset$;\\
   \STATE \textbf{return} MCS$(G_p,G_t, \alpha, curSol, maxSol)$;\\
  \end{algorithmic}
\end{algorithm}

McCreesh et al. \cite{Mccreesh2017partition} use an elegant way to represent $\alpha$, based on the fact that many vertices of $G_p$ have the same domain during the search. Thus the vertices of $G_p$ having the same $\alpha$ value should be put together to have a compact representation of $\alpha$. The following example illustrates this representation and its use in Algorithm \ref{alg:Search}.

\begin{figure}
	\centering
	\includegraphics[width=2.5in]{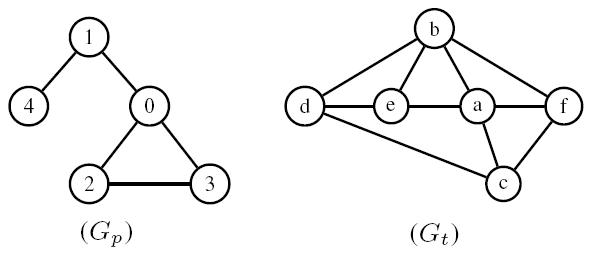}
	\vspace{-0.5em}
	\caption {An example for the MCS problem.}
	\label{fig1}
\end{figure}

\newtheorem{exam}{Example}
\begin{exam}
Figure \ref{fig1} shows two undirected and unlabelled graphs $G_p$ and $G_t$, $V_p$ = $\{0,1,2,3,4\}$, $V_t$ = $\{a,b,c,d,e,f\}$.
Initially, $\alpha(v) = \{a,b,c,d,e,f\}$ for each vertex $v$ of $G_p$, represented using the pair $\{\langle (0,1,2,3,4), (a,b,c,d,e,f)\rangle\}$.

Then, vertex 0 is chosen in line \ref{branchingH} for branching. The first match added into $curSol$ is $(0, a)$. Consequently,
$(1,2,3,4)$ is split into $(1, 2, 3)$ and $(4)$, and $(b,c,d,e,f)$ is split into $(b,c,e,f)$ and $(d)$, because vertices $1$, $2$ and $3$ are adjacent to the matched vertex $0$, while vertex $4$ is not; and vertices $b$, $c$, $e$ and $f$ are adjacent to the matched vertex $a$, while vertex $d$ is not. The updated $\alpha$ is then represented by $\{\langle (4), (d) \rangle, \langle (1,2,3), (b,c,e,f) \rangle\}$, saying that $\alpha(4) = \{d\}$ and $\alpha(1) = \alpha(2) = \alpha(3) = \{b, c, e, f\}$.

Note that the splitting of $(0,1,2,3,4)$ and $(a,b,c,d,e,f)$ is equivalent to removing $b$, $c$, $e$ and $f$
from $\alpha(4)$, and $d$ from $\alpha(1)$, $\alpha(2)$ and $\alpha(3)$.
\end{exam}

More generally, a pair $\langle V'_p, V'_t\rangle$ is called a label class in \cite{Mccreesh2017partition}, where $V'_p$ ($V'_t$) is a subset of $V_p$ ($V_t$), meaning that $\alpha(v) = V'_t$ for each $v \in V'_p$. Let $D$ be the set of label classes.  When a new match $(v, w)$ is added into $curSol$, Algorithm \ref{alg:Search} splits each label class $\langle V'_p, V'_t\rangle$ in $D$ into two label classes $\langle V'_{1p}, V'_{1t}\rangle$ and $\langle V'_{2p}, V'_{2t}\rangle$ in lines \ref{updateD1} -- \ref{updateD2}, so that the vertices in $V'_{1p}$ ($V'_{1t}$) are all adjacent to $v$ ($w$) and the vertices in $V'_{2p}$ ($V'_{2t}$) are all non-adjacent to $v$ ($w$). Note that $V'_{1p}$ and $V'_{2p}$, as well as $V'_{1t}$ and $V'_{2t}$, are disjoint.

This representation of $\alpha$ enables the following branching heuristic and bound computation in \cite{Mccreesh2017partition}.

\begin{itemize}
\item Given a label class $\langle V'_p, V'_t\rangle$, there are $|V'_p|\times|V'_t|$ matches to try. So, McCreesh et al. first select a label class such that $\max(|V'_p|, |V'_t|)$ is the smallest and then the vertex in $V'_p$ with the greatest degree in line \ref{branchingH} for branching, which is very similar to choosing a label class $\langle V'_p, V'_t\rangle$ with the smallest $|V'_p|\times|V'_t|$ and then breaking ties using vertex degrees. This heuristic is better than the heuristic used in \cite{NdiayeSolnon2011} consisting in selecting a label class with the smallest $|V'_t|$.
\item Let $D$ be the set of label classes at a branching point.
 A label class $\langle V'_p, V'_t\rangle$ can offer at most $\min(|V'_p|, |V'_t|)$ matches to $curSol$. So, the overestimate$(G_p, G_t, \alpha)$ function in \cite{Mccreesh2017partition} computes and returns $\sum_{\langle V'_p, V'_t\rangle \in D} \min(|V'_p|, |V'_t|)$, which is equivalent to the bound given by the soft all-different constraint in
\cite{NdiayeSolnon2011} but is much simpler to compute.
\end{itemize}

Nevertheless, the branching heuristic in \cite{Mccreesh2017partition} depends heavily on vertex degrees and may not result in the smallest search tree. In the next section, we will propose a new branching heuristic inspired by reinforcement learning.

\section{Learning Rewards for Branching}

In reinforcement learning, there is an agent and an uncertain environment. The agent is a learner and decision maker, and has a goal or goals. In order to achieve its goal, the agent interacts with the environment by taking actions and observing the impact of its actions to the environment.
Concretely, when the agent takes an action, it receives a reward related to its goal from the environment. It has a value function that transforms the cumulative rewards it received over time to a score of the action, representing a prediction of rewards in the future for this action. So, the agent should take the action with the maximum score among all available actions in each step to achieve its goal.  For a comprehensive presentation of reinforcement learning and its applications, see, e.g., \cite{reinforcementLearningIntroduction}.

Inspired by reinforcement learning, we regard Algorithm \ref{alg:Search} as an agent. It has a goal: reach a search tree leaf as early as possible to reduce as much as possible the search tree size. To achieve this goal, the agent adds successively a match $(v, w)$ into $curSol$. However,
it usually has many choices of $(v, w)$ in a step and does not know which choice is better. We then regard each choice $(v, w)$ as an action. Then, our key issue now is how to define the reward function and the value function.

As can be seen in Algorithm \ref{alg:Search}, the algorithm reaches a leaf when UB $\leq |maxSol|$. So, reducing UB quickly allows to reach a leaf quickly. Therefore, we define the reward $R$ for an action $(v, w)$ to be the decrease of UB after taking this action.
Concretely,
let $D$ be the set of label classes before taking the action $(v, w)$ and $D'$ the set of label classes obtained by splitting the label classes in $D$ according to their adjacency to $v$ and $w$ in lines \ref{updateD1} -- \ref{updateD2} of Algorithm \ref{alg:Search}, $R(v, w)$ can be quickly computed as follows.

\vspace{0.1cm}
$~~~~R(v, w) =  \sum_{\langle V'_p, V'_t\rangle \in D} \min(|V'_p|, |V'_t|) \ - $ \\
$~~~~~~~~~~~~~~~~~~~~~~~~~~~\sum_{\langle V''_p,  V''_t\rangle \in D'} \min(|V''_p|, |V''_t|) $ \\

Our value function maintains a score $S_p(v)$ ($S_t(w)$) for each vertex $v \in V_p$ ($w \in V_t$), initialized to 0. Each time $R(v, w)$ is computed, $S_p(v)$ and $S_t(w)$ are updated as follows:

\vspace{0.1cm}
$~~~~~~~~~~~~S_p(v) \leftarrow S_p(v) + R(v, w)$

$~~~~~~~~~~~~S_t(w) \leftarrow S_t(w) + R(v, w)$
\vspace{0.1cm}

At each branching point (line \ref{branchingH} of Algorithm \ref{alg:Search}), our algorithm first selects a label class $\langle V'_p, V'_t\rangle$ with the smallest $\max(|V'_p|, |V'_t|)$, and the vertex $v$ in $V'_p$ with the greatest score $S_p(v)$. Then, for each $w$ in $V'_t$ in the decreasing order of the score $S_t(w)$, the algorithm matches $v$ and $w$, and recursively continues the search after adding the match $(v, w)$ into $curSol$. All ties are broken in favor of the vertex with the maximum degree.

Note that original algorithm presented in \cite{Mccreesh2017partition} and our approach all select the label class
$\langle V'_p, V'_t\rangle$ with the smallest $\max(|V'_p|, |V'_t|)$. The difference lies in how to choosing $(v, w)$ from $\langle V'_p, V'_t\rangle$. In \cite{Mccreesh2017partition},  $v$ and $w$ are chosen according to their degree. In our approach, $v$ and $w$ are chosen according to their score. We will show the impact of this difference in MCS solving in the next section.

\section{Empirical Evaluation}
Experiments were performed on Intel Xeon CPUs E5-2680 v4@2.40 gigaHertz under Linux with 4G memory. The cutoff time is 1800 seconds for each instance.
We first present the algorithms and the benchmarks used in the experiments, and then present and analyze the experimental results.

\subsection{Solvers and Benchmark}
The following algorithms (also called solvers) are used in our experiments.

$\bullet$ McSplit~\cite{Mccreesh2017partition}:  An implementation of Algorithm \ref{alg:Search} using the label class representation of the $\alpha$ mapping.  It is more than an order of magnitude faster than the previous state of the art for unlabelled and undirected MCS instances. It is also extended for labelled graphs.

$\bullet$ McSplit$\downarrow$~\cite{Mccreesh2017partition}: A variant of McSplit using a top-down strategy to call the main McSplit method to search for a solution of cardinality $k=|V_p|, k-1, k-2, \ldots$ and  backtracks when the bound is strictly less than the required cardinality, and terminates when a solution of the required cardinality is found. This strategy is similar to the $k$$\downarrow$ algorithm~\cite{HoffmannEtAl2017}. McSplit$\downarrow$ is specially designed for the large subgraph isomorphism instances for which McSplit is beaten by $k$$\downarrow$.

$\bullet$ McSplit+RL: An implementation of Algorithm \ref{alg:Search} on top of McSplit with reinforcement learning presented in this paper. In other words, the only difference between McSplit+RL and McSplit is the use of reinforcement learning in the branching heuristic in McSplit+RL as presented in Section 4.

$\bullet$ McSplit+RL$\downarrow$: A variant of McSplit+RL using the top-down strategy of McSplit$\downarrow$.

The benchmark consists of two sets of instances.

$\bullet$ Biochemical reactions instances describing the biochemical reaction networks from the biomodels.net~\footnote{Available at http://liris.cnrs.fr/csolnon/SIP.html\label{bench}\\}. All the 136 graphs are directed, unlabelled bipartite graphs having 9 to 386 vertices. Every pair of graphs  
gives an MCS instance, resulting in 9316 {\em Bio} instances (including 136 self-match pairs).

$\bullet$ Large subgraph isomorphism and MCS instances~\cite{HoffmannEtAl2017,Mccreesh2017partition}. This benchmark set includes real-world graphs and graphs generated using random models, such as segmented images, modelling 3D objects, scale-free networks~\textsuperscript{\ref{bench}}. Pattern graphs range from 4 vertices to 900; target graphs range from 10 vertices to 6,671.
There are totally 12,427 instances, including: 6278 {\em images} instances from images-CVIU11 (there are 43 pattern graphs and 146 target graphs, each pair of pattern graph and target graph resulting in an instance); 1225 {\em LV} instances given by each pair of graphs (including two same graphs) among the 49 graphs selected in \cite{Mccreesh2017partition} from the LV set; 3430 {\em largerLV} instances from the above 49 LV graphs as pattern and the remaining 70 graphs as target in the LV set; 200 {\em Mesh}, 24 {\em PR15}, 100 {\em Scalefree} and 1170 {\em Si} instances used in \cite{Mccreesh2017partition} to evaluate McSplit.

\subsection{Performance of the new approach}

Table~\ref{tab:largergraphs} compares the performance of different algorithms for various instance sets. The first column gives the instance set name and the number of instances. The other columns gives the number of instances solved within the 1800s by the corresponding algorithm and the average runtime (inside parentheses) to solve these solved instances. McSplit+RL and McSplit+RL$\downarrow$ solve 130 and 117 instances more than McSplit and McSplit$\downarrow$, respectively. Note that the four solvers share the same implementation, and the only difference between McSplit and McSplit+RL, as well as between McSplit$\downarrow$ and McSplit+RL$\downarrow$, is their branching heuristic.

\begin{table*}[!htp]
\scriptsize
\caption{Comparison of the number of instances solved and the average running time (inside parentheses, in seconds) for large MCS instances.}
\label{tab:largergraphs}
\centering
\begin{tabular}{rrrrrrr}
\toprule
\emph{Instance set}($\# inst$) & \emph{McSplit} & \emph{McSplit+RL} & \emph{McSplit$\downarrow$} & \emph{McSplit+RL$\downarrow$}\\
\midrule		
Bio(9316) & 6655(45.8) & 6729(40.7) & 6818(43.2) & \textbf{6884(38.8)} \\	
Images(6278) & 1245(87.1) & 1280(87.9) & 1266(86.0) & \textbf{1283(95.6)}\\
LV(1225) & 400(59.4) & 418(64.8) & 410(44.4) & \textbf{425(39.1)} \\
LargerLV(3430) & 578(77.0) & 584(119.8) & 633(103.7) & \textbf{650(102.2)}\\
Mesh(200)  & -- & -- & \textbf{1(57.4)} & 1(754.8) \\
PR15(24) & 24(12.9) & 24(13.5) & \textbf{24(0.1)} & \textbf{24(0.1)}\\
Scalefree(100) & 13(0.0) & 13(0.0) & \textbf{80(7.3)} & 80(9.0)\\
Si(1170)  & 419(22.6) & 416(19.8) & 1157(3.7) & \textbf{1159(5.2)} \\
\midrule
total(21,743) & 9,334  &  9,464   & 10,389  & \textbf{10,506}  \\
\bottomrule
\end{tabular}
\end{table*}

\begin{figure*}[htp]
\begin{minipage}[t]{0.5\linewidth}
\centering
\includegraphics[width=3.5in,height=2.8in]{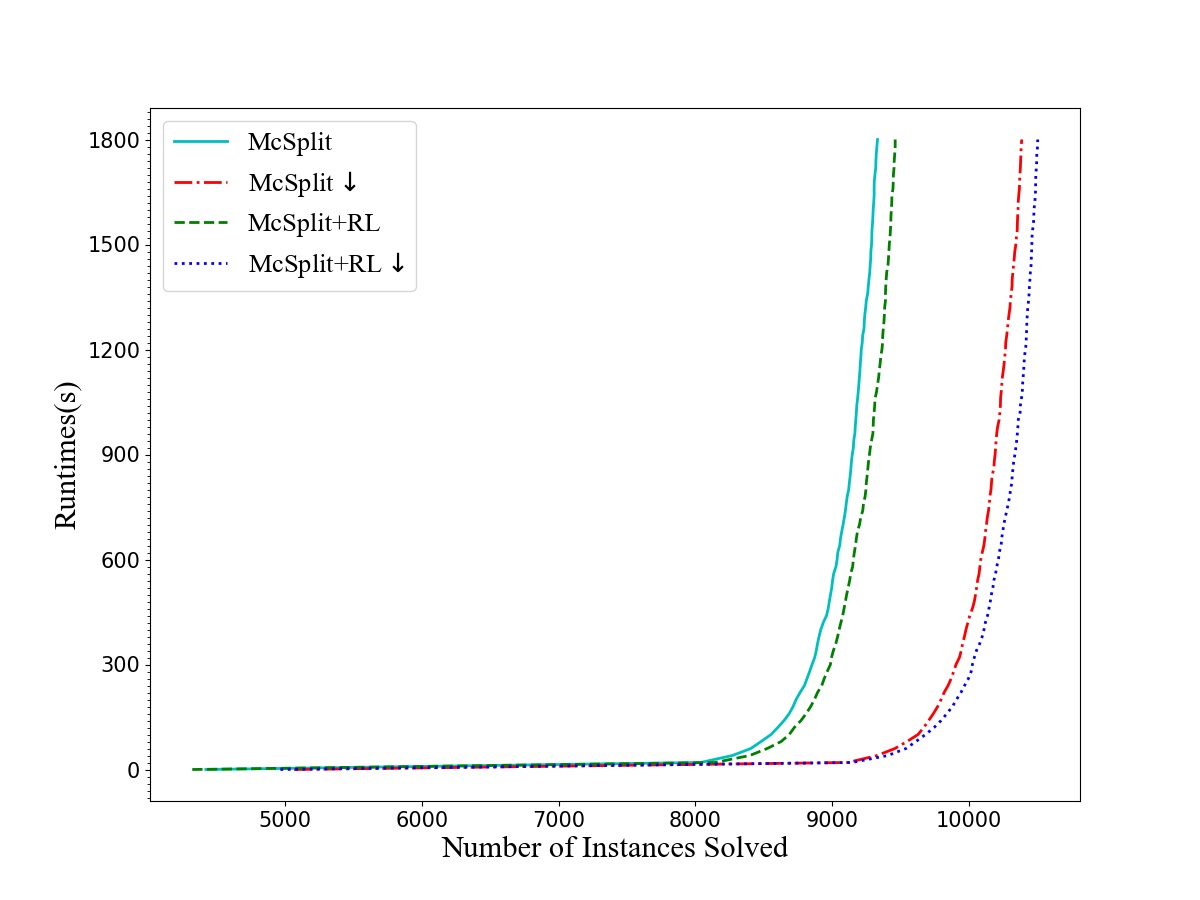}
\caption{Cactus plot of four solvers showing the number \protect\\ of instances solved over time among 21,743 instances}
\label{sumresult}
\end{minipage}%
\begin{minipage}[t]{0.5\linewidth}
\centering
\includegraphics[width=3.5in,height=2.8in]{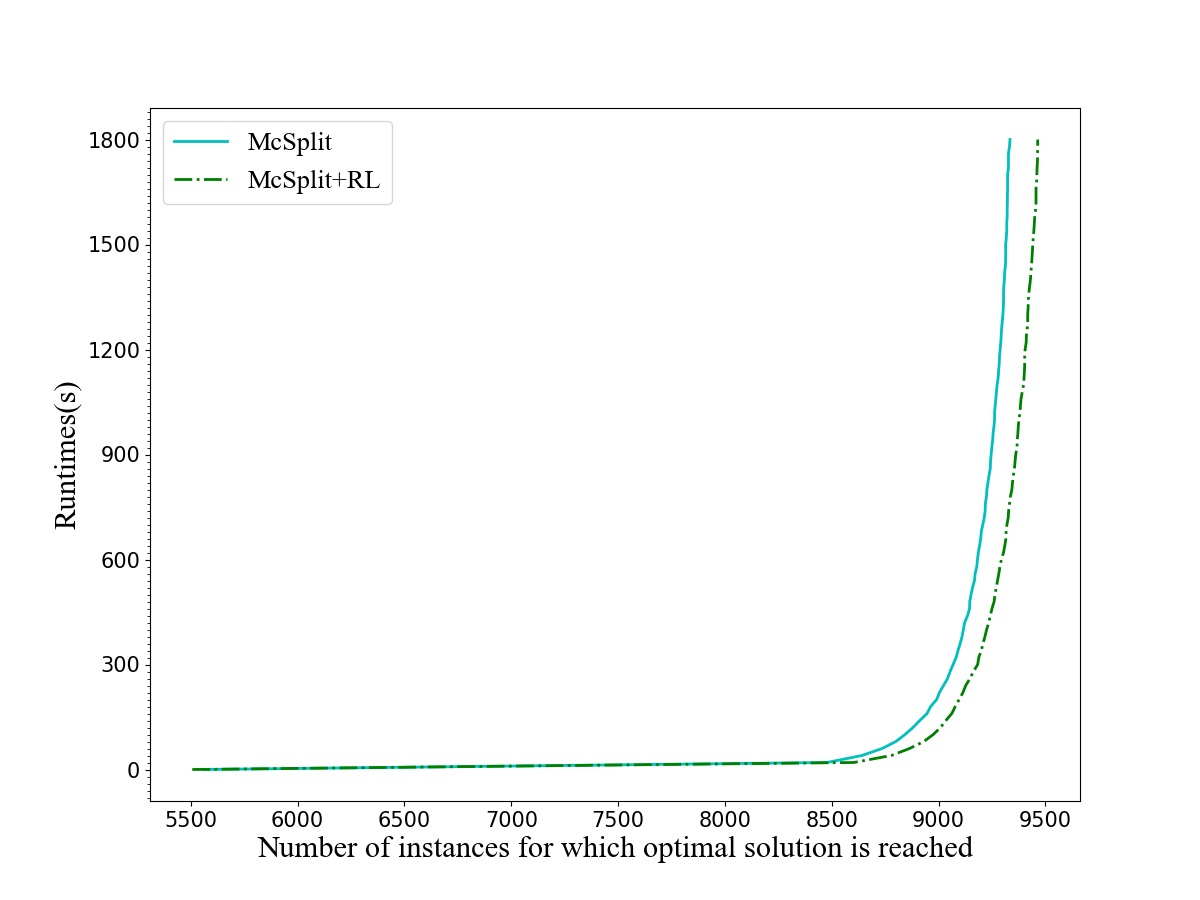}
\caption{Cactus plot of McSplit and McSplit+RL showing the \protect\\ number of instances for which the optimal solution is \protect\\ reached over time among 21,743 instances}
\label{bestresult}
\end{minipage}%
\end{figure*}

As Figure ~\ref{sumresult} showed, if we exclude the instances that are solved by both McSplit and McSplit+RL within 1s (5s, 10s), McSplit+RL solves 4.27$\%$ (6.26$\%$, 7.54$\%$) more instances than McSplit. The harder the instances are, the greater the effectiveness of reinforcement learning is. 

Recall that McSplit and McSplit$\downarrow$ are already highly efficient. The results show the effectiveness of the learning approach in designing branching heuristic in a BnB algorithm for the MCS, and the compatibility of the new branching heuristic with the top-down strategy of McSplit$\downarrow$.

\subsection{Analysis}

\begin{figure}[htp]
\centering
\includegraphics[width=3.5in,height=2.8in]{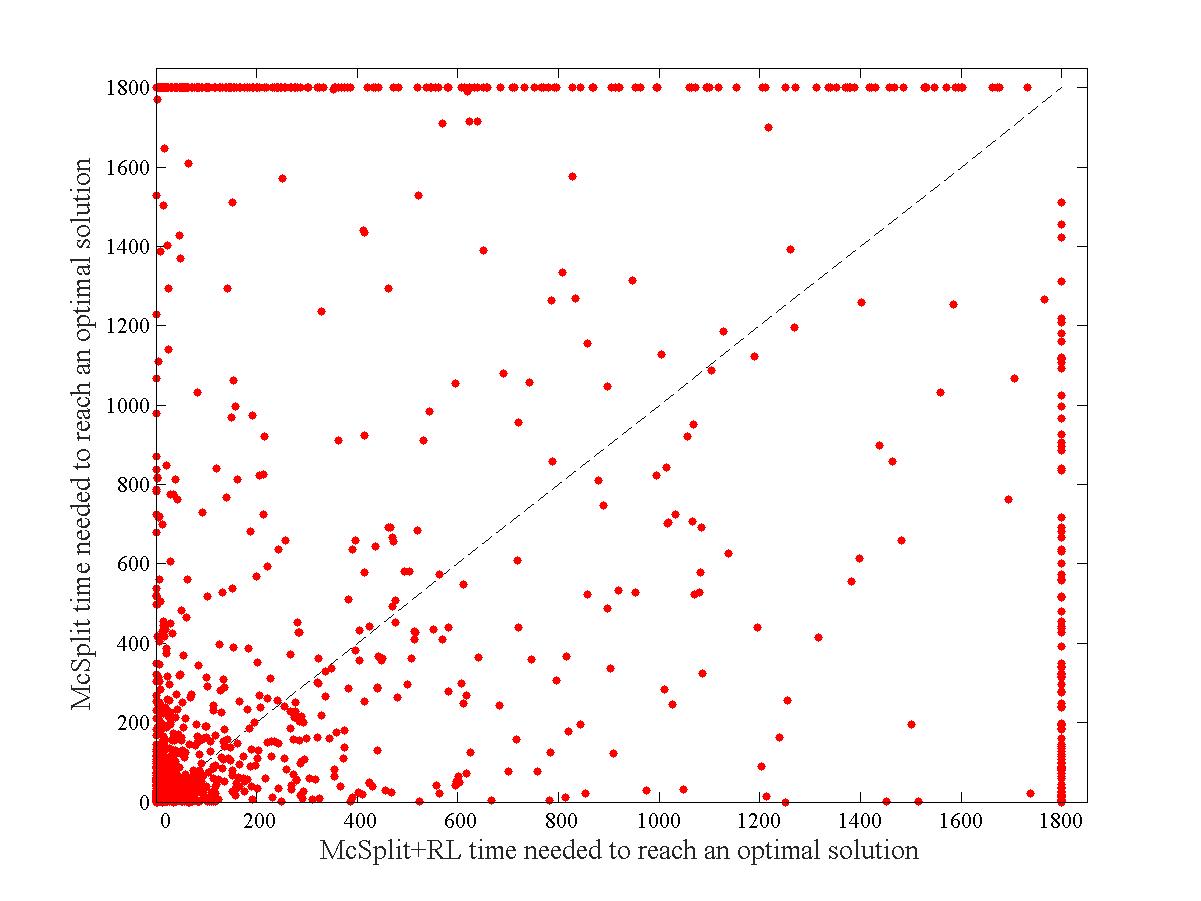}
\caption{Scatter plot comparing McSplit and McSplit+RL on the time needed to reach an optimal solution for 21,743 instances by excluding the easy instances solved by both McSplit and McSplit+RL within 10s and the too hard instances that none of McSplit and McSplit+RL solves within 1800s. There are 999 points above the diagonal line and 815 points under the diagonal line.}
\label{fig:side:a}
\end{figure}

The search of a BnB algorithm might be divided into two phases. In phase 1, the algorithm finds an optimal solution $s$; and in phase 2, it proves the optimality of $s$ by proving no better solution exists. Table \ref{tab2} shows, for a set of representative instances, the total solving time (phase 1 + phase 2) and the time for finding the optimal solution (phase 1) of McSplit and McSplit+RL, respectively. McSplit+RL usually finds the optimal solution more quickly than McSplit, allowing to prune the search more easily in line \ref{prune} of Algorithm \ref{alg:Search}. Note that when the time for finding the optimal solution is smaller, the time for solving an MCS instance usually is also smaller. As Figure ~\ref{fig:side:a} showed, McSplit+RL needs less time for finding the optimal solutions of 184 instances than McSplit.

\begin{table*}[!htp]
\caption{\small Comparison of the solving times ($time$),  the runtimes for finding the optimum ($time_{opt}$) and the standard deviation of the number of times selected for branching (in $10^5$) of vertex $v$ ($v_{sd}$) and $w$ ($w_{sd}$). $|V_p|$ ($|V_t|$) denotes the numbers of vertices of pattern (target) graphs. All times are in seconds. The best results are in bold.}
\centering
\scriptsize
\begin{tabular}{lcc|rrr|rrr}
\hline
\multirow{2}*{InstanceSet-$G_p$-$G_t$} & \multirow{2}*{$|V_p|$}  & \multirow{2}*{$|V_t|$} & \multicolumn{3}{c}{McSplit} & \multicolumn{3}{c}{McSplit+RL}  \\
\cline{4-9}
&  &  & $time$ & $time_{opt}$ & $v_{sd}$ ($w_{sd}$) & $time$ & $time_{opt}$ & $v_{sd}$ ($w_{sd}$)\\
\hline
 Bio-030.txt-061.txt & 50 & 73 & 372.30 & 371.60 & 623.74(204.04) & \textbf{184.65} & \textbf{183.30} & \textbf{187.68(50.20)} \\
 Bio-022.txt-046.txt & 38 & 31 & \textbf{238.65} & \textbf{32.42} & \textbf{309.68(79.55)} & 1119.95 & 247.70 & 1227.46(458.42) \\
 Bio-001.txt-018.txt & 46 & 79 & 1452.47 & 0.29 & 5214.98(545.06) & \textbf{905.52} & \textbf{0.09} & \textbf{1720.57(214.56)} \\
 Images-pattern11-target10 & 15 & 3506 & \textbf{0.15} & \textbf{0.07} & \textbf{0.02(0.00)} & 0.24 & 0.17 & 0.02(0.00) \\
 Images-pattern43-target113 & 89 & 2877 & 513.94 & 513.92 & 83.54(0.19) & \textbf{171.95} & \textbf{171.94} & \textbf{16.49(0.07)} \\
 Images-pattern24-target119 & 21 & 5376 & 17.51 & 15.48 & 2.81(0.00) & \textbf{9.64} & \textbf{7.20} & \textbf{0.95(0.00)} \\
 Images-pattern29-target120 & 22 & 4301 & 9.05 & 7.88 & 1.28(0.00) & \textbf{8.15} & \textbf{7.52} & \textbf{0.61(0.00)} \\
 LV-g10-g18 & 41 & 64 & 1103.46 & \textbf{854.83} & \textbf{1002.25(1.28)} & \textbf{1066.48} & 898.92 & 1419.91(1.78) \\
 LV-g12-g19 & 48 & 64 & 296.51 & 296.43 & 953.03(180.30) & \textbf{18.60} & \textbf{7.85} & \textbf{13.22(5.98)} \\
 LV-g11-g21 & 42 & 64 & \textbf{2.06} & 0.00 & 4.10(1.00) & 3.11 & 0.00 & \textbf{3.26(0.99)} \\
 LV-g10-g17 & 41 & 64 & 203.83 & 158.45 & \textbf{241.36(20.52)} & \textbf{149.38} & \textbf{131.21} & 271.18(5.43) \\
 LargerLV-g11-g78 & 42 & 627 & 6.23 & 6.23 & 2.33(0.27) & \textbf{2.13} & \textbf{2.13} & \textbf{0.43(0.07)} \\
 LargerLV-g12-g55 & 48 & 256 & 1471.98 & 1423.98 & 3580.15(73.53) & \textbf{562.62} & \textbf{246.93} & \textbf{345.12(30.99)} \\
 LargerLV-g13-g70 & 49 & 501 & \textbf{213.09} & 213.08 & 234.17(0.61) & 401.03 & \textbf{197.28} & \textbf{187.58(1.66)} \\
 LargerLV-g6-g72 & 19 & 561 & 18.46 & 18.46 & 33.90(0.09) & \textbf{0.02} & \textbf{0.02} & \textbf{0.02(0.00)} \\
 LargerLV-g6-g71 & 19 & 501 & \textbf{0.01} & \textbf{0.01} & \textbf{0.02(0.00)} & 0.14 & 0.02 & 0.14(0.00) \\
 PR15-pattern1-target & 83 & 4838 & \textbf{9.74} & \textbf{9.68} & \textbf{0.27(0.02)} & 23.31 & 23.25 & 0.54(0.02) \\
 PR15-pattern9-target & 68 & 4838 & 5.56 & 5.49 & 0.24(0.00) & \textbf{5.09} & \textbf{4.99} & \textbf{0.16(0.00)} \\
 Si-si2\_b03m\_m200.05 & 40 & 200 & 2.08 & 2.08 & 3.19(0.94) & \textbf{0.09} & \textbf{0.09} & \textbf{0.16(0.03)} \\
 Si-si2\_m4Dr2\_m256.02 & 51 & 256 & 1516.22 & 1516.22 & 1169.12(127.65) & \textbf{1156.89} & \textbf{1156.89} & \textbf{460.27(70.89)} \\
\noalign{\smallskip}\hline
\end{tabular}
\label{tab2}
\end{table*}
Let $b_p(v)$ ($b_t(w)$) denote the number of times a vertex $v$ in the pattern graph  $G_p$ ($w$ in the target graph $G_t$) is used for branching at line \ref{branchingH} (line \ref{branchingH1}) of Algorithm \ref{alg:Search}. Table \ref{tab2} also gives the standard deviation $v_{sd}$ ($w_{sd}$) of $b_p(v)$ ($b_t(w)$) when McSplit and McSplit+RL solves these representative graphs. This standard deviation with McSplit+RL is usually significantly smaller than with McSplit, meaning that more vertices participate in branching in McSplit+RL than in McSplit.

This phenomenon might be explained as follows. McSplit always branches on the vertex with the maximum degree given a label class, concentrating the branching on a small subset of vertices with high degree. However, McSplit+RL can also branch on vertices with lower degree in a label class, because reinforcement learning discovers that these vertices allow a big bound decrease.

In summary, the search of McSplit+RL is more diversified while leading to quick pruning, explaining why McSplit+RL usually finds the optimal solution more quickly than McSplit.

\section{Conclusion}
We proposed a branching heuristic inspired from reinforcement learning in a BnB algorithm for the MCS, by regarding the algorithm as an agent and a match $(v, w)$ as an action. The reward of an action is the decrease of the upper bound and the score of a vertex is the sum of rewards of the actions it participated in the past. Then, the algorithm uses these scores to select the action $(v, w)$ for branching. Intensive experiments show that this branching heuristic allows to solve significantly more instances, because it allows a more diversified search.

Our results suggest that reinforcement learning is a very promising tool for NP-hard problem solving. We will improve the   reward and value function definitions to further improve our branching heuristic for the MCS in the future.

\bibliographystyle{named}
\small
\bibliography{ijcai19MCSP}

\end{document}